%% file: paper.tex
\documentclass[sigconf]{acmart}
\AtBeginDocument{%
  \providecommand\BibTeX{{%
    \normalfont B\kern-0.5em{\scshape i\kern-0.25em b}\kern-0.8em\TeX}}}

\setcopyright{acmlicensed}
\acmJournal{PACMMOD}
\acmYear{2025} \acmVolume{3} \acmNumber{3 (SIGMOD)} \acmArticle{119} \acmMonth{6}\acmDOI{10.1145/3725256}

\acmConference[SIGMOD '25]{ International Conference on Management of Data }{June 22-27, 2025}{Berlin, Germany}
\usepackage[most]{tcolorbox}
\usepackage[framemethod=tikz]{mdframed} % Colored box
\usepackage{fancyvrb}
\usepackage{xspace}
\usepackage{appendix}
\usepackage{enumitem}
\usepackage{subfigure}

\newcommand{\ourapproach}{\emph{LTS}\xspace}

\newcommand{\review}[1]{\textcolor{blue}{#1}\xspace}
\renewcommand{\review}[1]{#1\xspace}

\def\withnotes{1} % change to (0) to hide all comments, (1) to show all comments

\definecolor{Maroon}{rgb}{0.62, 0.0, 0.09}
\definecolor{Emerald}{rgb}{.07, .74, .62}

\ifnum\withnotes=1
  \newcommand{\jfcolor}[1]{{\color{Emerald}#1}} % Juliana
  \newcommand{\jbcolor}[1]{{\color{violet}#1}} % Juliana Barbosa
  
  \newcommand{\ascolor}[1]{{\color{violet}#1}} % Aecio
  
  \newcommand{\asnote}[1]{\ascolor{\textbf{AS: }\sf #1}}
  \newcommand{\as}[1]{\ascolor{\textbf{Aécio: }\sf #1}}
  \newcommand{\jb}[1]{\jbcolor{\textbf{JB: }\sf #1}}
  \newcommand{\jf}[1]{\jfcolor{\textbf{JF: }\sf #1}}
  \else
  \newcommand{\asnote}[1]{}
  \newcommand{\as}[1]{}
  \newcommand{\jb}[1]{}
  \newcommand{\jf}[1]{}
  
\fi

\newcommand{\hide}[1]{}
\newcommand{\myparagraph}[1]{\vspace{0.25em}\noindent \textbf{#1.}}

\newcommand{\myparagraphemph}[1]{\vspace{0.25em}\noindent \emph{#1.}}

\renewcommand{\paragraph}[1]{\vspace{0.1em}\noindent \textit{#1.}}

\usepackage{amsmath}
\usepackage{algorithm}
\usepackage{algpseudocode}
\usepackage[table]{xcolor}

\usepackage{graphicx}
\usepackage{subcaption}
\usepackage{colortbl}
\usepackage{multirow}
\usepackage{caption}
\usepackage{booktabs}

\newenvironment{myitemize}%
{\begin{list}{$\bullet$}{%
			\setlength{\labelsep}{2pt}\setlength{\leftmargin}{5pt}%
			\setlength{\labelwidth}{0pt}%
			\setlength{\listparindent}{0pt}}}
{\end{list}}

{\begin{list}{\noindent}{%
	\setlength{\leftmargin}{0pt}%
	\setlength{\listparindent}{0pt}}}
{\end{list}}
\begin{document}

%\title{Leveraging GPT-Generated Labels to Enhance Wildlife e-Commerce Advertisement Detection}
%\title{A Cost-Effective LLM-based Approach to Identify Wildlife Advertisements in Online Marketplaces}
%\title{A Low-Cost LLM-based Approach to Identify Wildlife Advertisements}
%\title{Using LLMs to Identify Wildlife Advertisements}
\title{A Cost-Effective LLM-based Approach to Identify Wildlife Trafficking in Online Marketplaces }
% \\ {\it (Data-Driven Applications)}}

\author{Juliana Barbosa$^1$ \ \ Ulhas Gondhali$^2$ \ \ Gohar Petrossian$^2$ \ \ Kinshuk Sharma$^3$ \ \ Sunandan Chakraborty$^3$ \ \ Jennifer Jacquet$^4$ \ \ Juliana Freire$^1$}

%\email{juliana.barbosa@nyu.edu, ugondhali@jjay.cuny.edu, gpetrossian@jjay.cuny.edu, kisharma@iu.edu,sunchak@iu.edu,jjacquet@miami.edu, juliana.freire@nyu.edu}
\email{{juliana.barbosa,juliana.freire}@nyu.edu, {ugondhali,gpetrossian}@jjay.cuny.edu,} 
\email{{kisharma,sunchak}@iu.edu,jjacquet@miami.edu}

\affiliation{%
  \institution{$^1$New York University \ \ $^2$John Jay College of Criminal Justice \ \ $^3$Indiana University Indianapolis \ \ $^4$University of Miami}
  \country{}
  \city{}
  \state{}
  
}

\renewcommand{\shortauthors}{Juliana Barbosa et al.}

\begin{abstract}
\input{abstract}

\end{abstract}

\begin{CCSXML}
<ccs2012>
<concept>
<concept_id>10010147.10010257.10010282.10011304</concept_id>
<concept_desc>Computing methodologies~Active learning settings</concept_desc>
<concept_significance>500</concept_significance>
</concept>
</ccs2012>
\end{CCSXML}

\ccsdesc[500]{Computing methodologies~Active learning settings}


% \keywords{Do, Not, Us, This, Code, Put, the, Correct, Terms, for,

%      \vspace{5em}


\maketitle

\section{Introduction}
\label{sec:intro}
\input{introduction}

\section{Related Work}
%\review{MOVE RELATE WORK TO BEGINNING}
\label{sec:related-work}
\input{related-work}

\section{Using LLMs and Active Learning to Label Wildlife Ads}
\label{sec:training-data-generation}
\input{training-data-generation}

\section{Experimental Evaluation}
\label{sec:experimental-evaluation}
\input{experimental-evaluation}

\vspace{.3cm}
\section{Use Cases}
\label{sec:use-cases}
\input{use-cases}

\section{Conclusion}
\label{sec:conclusion}
\input{conclusion}

\myparagraph{Acknowledgments}
We thank Spencer Roberts for contributing the analysis of shark trade.
This work was supported by NSF awards CMMI-2146306, CMMI-2146312,
CMMI-2146306,  and IIS-2106888.
Freire was partially supported by the  DARPA Automating Scientific Knowledge
Extraction and Modeling (ASKEM) program, Agreement No. HR0011262087.  The views, opinions, and findings expressed are those of the authors and should not be interpreted as representing the official views or policies of the DARPA, the U.S. Government, or NSF.

%% IIS 2106888 Dataset Search and Discovery
%% CMMI 2146306 Wildlife NYU; 2146312 John Jay
%% OAC 2411221 Urban
% \vspace{3cm}
\balance
\bibliographystyle{ACM-Reference-Format}
\bibliography{paper}

\end{document}

%% file: abstract.tex
Wildlife trafficking remains a critical global issue, significantly impacting biodiversity, ecological stability, and public health. Despite efforts to combat this illicit trade, the rise of e-commerce platforms has made it easier~to sell wildlife products, putting new pressure on wild populations of endangered and threatened species.
The use of these platforms also opens a new opportunity: as criminals sell wildlife products online, they leave digital traces of their activity that can provide insights into trafficking activities as well as how they can be disrupted.
The challenge lies in finding these traces. Online marketplaces publish 
ads for a plethora of products, and identifying ads for wildlife-related products is like finding a needle in a haystack. 
Learning classifiers can automate ad identification, but creating them requires costly, time-consuming data labeling that hinders support for diverse ads and research questions.
This paper addresses a critical challenge in the data science pipeline for wildlife trafficking analytics: generating quality labeled data for classifiers that select relevant data. While large language models (LLMs) can directly label advertisements, doing so at scale is prohibitively expensive. We propose a cost-effective strategy that leverages LLMs to generate pseudo labels for a small sample of the data and uses these labels to create specialized classification models. Our novel method  automatically gathers diverse and representative samples to be labeled while minimizing the labeling costs. Our experimental evaluation shows that our classifiers achieve up to 95\% F1 score, outperforming LLMs at a lower cost. We  present real use cases that demonstrate the effectiveness of our approach in enabling analyses of different aspects of wildlife trafficking.

%% file: introduction.tex
Wildlife trafficking is a global issue that affects the environment with biodiversity loss, ecological disruption, and health concerns \cite{mozer2023introduction, demeau2019wildlife, scheffers2019global}. It is one of the most profitable illegal commerce in the world---generating between \$7-\$23 billion dollars per year \cite{citesreport2022}---and encompasses multiple market segments including fashion, exotic pets, traditional medicine, wild food, and accessories \cite{mozer2023introduction}. 

Despite the increasing efforts in combating wildlife trafficking, some suggest that during the COVID-19 pandemic trafficking escalated as traders were forced to shift from face-to-face to online interactions~\cite{traffic-covid-investigation-2021,traffic-covid-eia-2020}. 
This, combined with the proliferation of online marketplaces which provide a convenient mechanism for globalized buying and selling~\cite{nalluri2021survey}, has put new pressure on wild populations of endangered and threatened species~\cite{vietnamese2016rapid,xu2020illegal,haysom2019search}.
Notwithstanding, the move to cyberspace brings an opportunity: as criminals sell wildlife products, they leave digital traces of their activity and, by analyzing these traces, we can obtain insights into trafficking activities as well as how they can be disrupted.

The challenge lies in finding these traces. Online marketplaces publish a very large number of ads and identifying ads for wildlife-related products is like finding a needle in a haystack.
To retrieve ads, we need to issue queries to marketplaces that invariably return  irrelevant results. 
Consider for example an expert trying to assess the volume of the trade for endangered shark species.
If we search for \textit{shark} on ebay~\cite{ebay}, we will find ads for toys, shirts, plates, vacuum cleaners (of the Shark brand), among products that actually contain shark body parts.
Even a more specific query such as \textit{shark jaw} returns many irrelevant ads, e.g.,
shark teeth fossils which are unrelated to wildlife trafficking.
Therefore, to produce trustworthy results, experts must curate the collected data and ensure that the ads they use in the analysis are indeed for animal-derived products.

Since assessing the relevance of a large number of ads requires a time-consuming process, the wealth of marketplace data has been largely untapped:
studies on online wildlife trafficking are often limited to short time spans for specific species~\cite{hernandez2015automatic, siriwat2018illegal, di2018machine,harrington2019popularity, martin2018trade,gomez2019bearly,venturini2020disguising,roberts2022systematic} or trade in a specific region~\cite{vietnamese2016rapid,roberts2022systematic,stoner2014tigers}.
This has significantly hampered a more comprehensive understanding of wildlife traffic~\cite{haas2015federated, charity2020wildlife}.

\begin{figure*}[t!]
\centering
\includegraphics[width=1\textwidth]{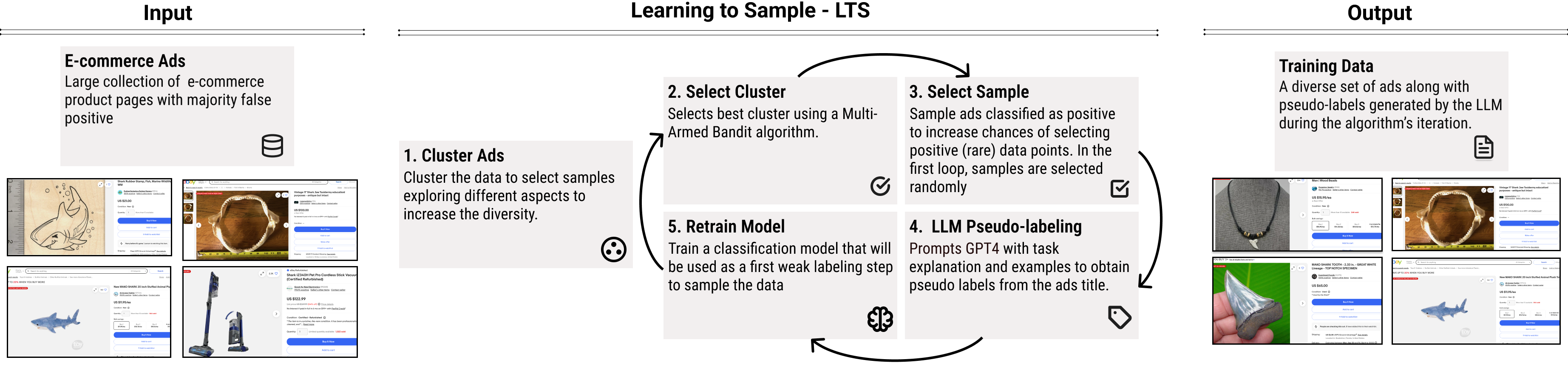} % Adjust the width here
\caption{\ourapproach Overview: Given a large collection of ads obtained from online marketplaces and a natural language description of the ad selection criteria for a given research question,  \ourapproach selects a representative sample of the ads and leverages LLMs to label them. The labeled data can then be used to create classifiers specialized for the research question.
}
\Description{This figure illustrates the LTS approach. From input of ads to the final data labeled collection.}
\label{fig:csb-data-generator}
\end{figure*}

\myparagraph{Learning-Based Approaches to Identify Wildlife-Related Ads}
One possible approach to tackle this challenge is to use learning-based classifiers. However, obtaining data to train learning-based models is costly. In addition, wildlife trafficking has many facets and different research questions demand different data slices, which in turn require specialized classifiers.

\begin{example}[Animal Products] \label{example:animal}
\review{Wildlife trafficking is a global problem involving many species and product types. A broad search must be performed over animal-derived products advertised online to understand the landscape of wildlife trafficking.}
\end{example}
%\begin{tcolorbox}[colback=green!5!white,colframe=green!75!black]
\begin{example}[Small Leather Products] \label{example:leather}
Within the broader context of wildlife trafficking, the illegal trade of small leather products poses a significant concern, as studies reveal that these items are among the most commonly trafficked illegal wildlife products globally \cite{petrossian2016overview, van2019comparison}. 
To study the impact of this market on endangered species such as alligators and snakes, experts need to obtain ads for a plethora of products such as belts, wallets, and bags, made with the skin of these animals. 
\end{example}

\begin{example}[Shark Products] \label{example:shark}
Shark conservation represents a critical global diversity issue, with many shark populations experiencing significant decline due to fishing practices \cite{pytka2023internet}.
Recreational land-based shark fishing is on the rise globally, raising concerns about its impact on threatened shark populations \cite{hingley2020conservation}.
Experts studying shark trade are interested in exploring products  associated with animals that were killed for their parts
(e.g., shark teeth, jaws, fins). At the same time, their analysis should not include fossils (e.g., fossil teeth) which do not impact the shark population.  
\end{example}

For each of these tasks, experts must label a representative and a sufficiently large set of ads to create an effective classifier.
Given the stark difference between the species and products under study, often,
they are not able to re-use classifiers--a classifier built to identify small leather products is not effective for classifying shark-related products or to determine whether a product is derived from an animal (and vice-versa).
The need to manually label a large number of ads for each task greatly limits the experts' ability to explore multiple research questions.

Foundation models offer an effective alternative to perform classification in a zero-shot fashion given that they embody a large amount of knowledge and have been shown to \textit{understand} natural language~\cite{touvron2023llama, radford2018improving,brown2020language}. 
Their use can greatly reduce the need for human input and make it possible to automatically curate data for different research questions. However, doing so can be expensive given the large number of ads that need to be classified. As a point of reference, using GPT-4 to label 800,000 ads collected over one week would cost over \$17,000. 
The cost makes this alternative impractical and out of reach for most researchers and under-resourced institutions~\cite{wcs_about}. Open-source models like LLAMA3 (text) and LLAVA (text and image) present a promising solution to mitigate cost burdens. However, as demonstrated in our experiments (see Table \ref{table:llms_baseline_combined}), these models have much lower accuracy.
Another important aspect that should not be neglected is the environmental impact of LLMs due to their high energy consumption~\cite{rillig2023risks}; given the small percentage of relevant ads, their use leads both to energy waste and a very high cost per \textit{useful} ad.

\myparagraph{Using LLMs as Annotators}
To support the effective classification of wildlife-related ads for different research questions (and  collection tasks) at a low cost, 
we propose  \ourapproach (Learn to Sample).
A key insight behind \ourapproach is to leverage in-context learning (few-shot) \cite{brown2020language} with prompt-based text-generation techniques over \emph{a subset of the data collected}. These labels can then serve as pseudo-ground truths for training smaller, specialized classifiers, thus substantially reducing the number of inferences (and cost) required for the LLMs. But selecting this subset set is difficult. 

To create classifiers that are effective and robust, we need to obtain a diverse, representative set of positive and negative samples. Candidate ads for wildlife products collected from e-commerce sites are inherently unbalanced--very few of the retrieved ads are actually related to living animals. A recent study that
%\jb{we need to put in third person} 
analyzed roughly 1 million ads and found that fewer than 51 thousand ads were relevant for the wildlife domain \cite{barbosa2024flexiblescalableapproachcollecting}.
Therefore, labeling random samples of the data is unlikely to result in good classifiers.  
We propose a new method that learns to select a set of representative samples. As illustrated in Figure~\ref{fig:csb-data-generator}, \ourapproach (Learn to Sample) clusters the data to ensure that a diverse set of ads is selected. In addition, it uses a multi-armed bandit-based strategy to balance exploration and exploitation with the goal of selecting a sufficient number of positive examples.
Through an iterative process, \ourapproach selects samples from different clusters and applies active learning, leveraging LLMs as labelers. This strategy enables the construction and incremental refinement of a model for classifying and selecting relevant ads for inclusion in the sample.

We experimentally evaluated the effectiveness of \ourapproach (Section~\ref{sec:experimental-evaluation}) 
using both text-based and multi-modal (MM) classification methods.
Our results show that models trained with LTS-labeled data achieve high accuracy, outperforming those trained with random or manually labeled samples. When compared to zero-shot LLM classifiers, the LTS-derived models substantially outperform open-source models and perform better than (or comparably to) GPT-4, despite using only a few million parameters instead of GPT-4's trillions, at a fraction of the cost.

\myparagraph{Contributions}
Our main contributions can be summarized as follows:
\begin{myitemize}
\item   
We propose \ourapproach, a cost-effective approach that leverages LLMs to label wildlife by combining clustering with a multi-armed bandit algorithm to obtain diverse, representative samples, and applying active learning to reduce  labeling costs (Section~\ref{sec:training-data-generation}).

\item  We conduct an experimental evaluation (Section~\ref{sec:experimental-evaluation}) using three (real) research questions with varying data collection requirements, each presenting different collection size (from 15k to 700k ads) and complexity levels (from specific to generic questions). The results show that models trained on LTS-labeled data consistently achieve high accuracy.  
\item We study how the performance of \ourapproach varies using five different LLMs for labeling and 
examine its scalability by comparing the runtime performance over ad collections of different sizes. 
\item We present two real use cases carried out by environmental scientists and criminologists using data derived by \ourapproach. They illustrate the potential of our approach to enabling experts to explore research questions that can lead to valuable insights into the dynamics of illicit wildlife trade. 
\end{myitemize}
Our code is open-source and available at \url{https://github.com/VIDA-NYU/LTS}.

%% file: related-work.tex
% Task 1
\paragraph{Labeling and Sampling Data to Create Classifiers} 
While learning-based models have been shown effective for a wide range of classification tasks, gathering sufficient high-quality data to train these models is a major bottleneck.  This problem has received attention in the database community and approaches have been proposed to reduce human effort in labeling training data~\cite{das2020goggles, whang2023data, heo2020inspector, varma2018snuba, ratner2017snorkel}. Snorkel introduced a data programming approach to labeling in which users can generate training data by writing heuristic labeling functions~\cite{ratner2017snorkel}. To reduce the user effort, Snuba proposed the use a small labeled set to automatically generate heuristics~\cite{varma2018snuba}. Other approaches have focused on image data. Inspector Gadget combined data programming, crowdsourcing, and data augmentation to generate weak labels for image classification~\cite{heo2020inspector}, while Goggles used affinity functions to label image datasets~\cite{das2020goggles}. For a detailed survey on labeling approaches, see~\citet{whang2023data}. 
We propose to leverage LLMs for labeling data at scale. Given their broad knowledge, LLMs can serve as general purpose classifiers, but using them for large scale data triage tasks, such as identifying wildlife-related ads, can be prohibitively expensive.  LTS introduces a cost-effective approach that optimizes the use of LLMs: it uses expensive LLMs sparingly (as a weak labeler) and trains smaller/cheaper models to be efficiently applied at scale.  
In contrast to prior work, LTS does not require users to write labeling functions or provide labeled data; to customize a labeling task, users specify a prompt that defines the classification task.

\paragraph{Online Wildlife Trafficking Detection}
Our work is related to wildlife trafficking detection, with a focus on automated techniques and natural language processing (NLP) applications.
Keskin et al.  \cite{keskin2022quantitative} offers a comprehensive perspective on the illicit wildlife trade, shedding light on the critical challenges encountered in this domain. One of the most significant challenges highlighted is the scarcity of data, exacerbated by the fact that the available data tends to be biased toward specific regions and particular species. For instance, Cardoso et al.~\cite{cardoso2023detecting} trained a neural network to identify pangolins and Qing Xu~\cite{xu2019use} used Twitter data to detect wildlife trade,  focusing on one animal species (Pangolins) and one animal product (Ivory).

Kulkarni and Di Minin \cite{kulkarni2023towards} highlight the usefulness of machine learning to power approaches to wildlife conservation but also discuss the challenges in finding high-quality training data. In their study, a novel dataset of images was created to help distinguish animals in captivity and in the wild. 
The field of wildlife image detection has witnessed extensive research efforts \cite{xu2019use, cardoso2023detecting, kulkarni2023towards} with the development of large-scale datasets coming from different sources, such as camera traps \cite{meng2023method, lightweight} or web-harvested images \cite{chabot2022using, roy2023wildect}. Beyond image detection, approaches have applied deep learning for text classification within the wildlife context to identify sources documenting the hunting of bats~\cite{hunter2023using} and bird trading in Australia~\cite{stringham2021text}.
For a broader perspective on machine learning for wildlife conservation see Tuia et al.~\cite{tuia2022perspectives}. 
These studies underscore the effectiveness of machine learning methods to automatically identify wildlife trade, but they also highlight important challenges: the need for large volumes of training data as well as the inability of these models to generalize for different species/products. 
Our approach aims to address these challenges. 

\paragraph{Using LLMs as Evaluators and Annotators}
LLMs are increasingly being used as an alternative to human evaluators for different tasks. %from annotation to evaluation. 
Approaches have been proposed to use LLMs to evaluate
natural language generation (NLG) systems as well as for tasks that have traditionally relied on human judgement~\cite{evalullm@iui2024,liu-etal-geval-2023,ye-flask@arxiv2023,chiang-lee-large@acl2023}. LLM-based methods have also been used as general classifiers for different tasks, including column-type annotation, schema and data matching~\cite{chorus@vldb2024,unicorn@sigrec2024}.
A key concern in our application scenario is the sheer number of ads to be labeled and the associated cost. Our approach thus leverages LLMs to generate training data to support the creation of more cost-effective classifiers that can be specialized for different tasks.

\paragraph{Active Learning for Imbalanced Data}
Dealing with class imbalance, where the number of examples in each class differs significantly, is a common challenge that has been extensively studied in the machine learning literature~\cite{rezvani2023broad,volk2024adaptive,chawla2002smote,lin2017clustering,salehi2024cluster,chamlal2024hybrid}. Given a set of labeled instances, these strategies select a subset to be used for training, for example, through random sampling, over or under-sampling \cite{gosain2017handling,chawla2002smote}, or applying clustering algorithms \cite{lin2017clustering}.
As discussed in Section~\ref{sec:intro}, wildlife-related ads are much less frequent than irrelevant ads, but since we do not know their labels a-priori, we cannot directly apply existing methods to handle class imbalance.

When unlabeled data is abundant and labels are expensive to obtain, \textit{active learning} (AL) methods can be used to select the most informative data points to label, rather than labeling the entire dataset. AL methods iteratively select data points and query an oracle to obtain the labels for the points; they aim to minimize the number of queries while maximizing the performance of the resulting classification model~\cite{settles2009active,wang2011active}.  
Different selection strategies are possible, including: 
\textit{uncertainty sampling} selects instances for which the oracle is most uncertain about~\cite{pmlr-v162-raj22a, yang2015multi,monarch2021human}, and
\textit{diversity sampling} aims to obtain a representative sample by selecting instances that cover a wide range of features or classes~\cite{tharwat2023survey}.
\ourapproach (Section~\ref{subsec:proposed}) combines elements of methods for handling class imbalance and active learning. It uses an LLM and a model that is incrementally refined as Oracles and applies uncertainty and diversity sampling to select ads to label.

Approaches for active learning that address class imbalance have been proposed in the literature~\cite{lesci2024anchoral,coleman2022similarity,aggarwal2021minority}.
AnchorAL~\cite{lesci2024anchoral} and SEALS~\cite{coleman2022similarity} promote the discovery of rare instances and improve the computational efficiency of AL by selecting instances that are most similar to the labeled examples, instead of relying on more expensive model inferences.
\citet{aggarwal2021minority} proposed a minority class-oriented sample acquisition, which similar to our work, is based on the samples predicted by the latest model learned. They apply the core-set \cite{sener2017active} method to select a diverse sample. To obtain a diverse sample, our approach clusters the unlabeled instances, and to promote the discovery of the rare positive examples, it applies a multi-armed-bandit-based algorithm that selects 
instances from the different clusters. 
Given the size of our data, using model inference has proven to be effective: as discussed in Section~\ref{sec:experimental-evaluation}, our approach derives high-quality classifiers at a low cost.
Using similarity measures for selecting instances as well as core-sets for attaining diversity are directions we plan to pursue in future work.

\paragraph{Labeled Data Acquisition}
The idea of using a combination of clustering and multi-armed bandits (MAB) has been applied to a different problem: labeled data acquisition~\cite{chai2022selective,wang2024optimizing}. Given a model and associated training data, they aim to find, among a collection of datasets, tuples to augment the training data that improve the model. 
Unlike our scenario, in this problem the labels are available and they do not have to deal  with class imbalance.

%% file: training-data-generation.tex
To construct machine learning-based classifiers that are able to accurately filter ads that are
relevant for a research question, we need to collect and label a large number of ads.
Our approach makes use of LLMs to automate the labeling process and introduces a new strategy to create a set of representative samples. 

\subsection{Wildlife Data Curation Pipeline}
\label{sec:approach-overview}
\input{approach-overview}

\subsection{Labeling Ads Automatically} While the cost of using GPT-4 can be prohibitive for a large number of ads, doing so for a small sample is cost-effective.
We make use of in-context learning (few-shot) to derive class labels for ads.

\begin{figure*}[t]
    \centering   
    \includegraphics[width=\textwidth]{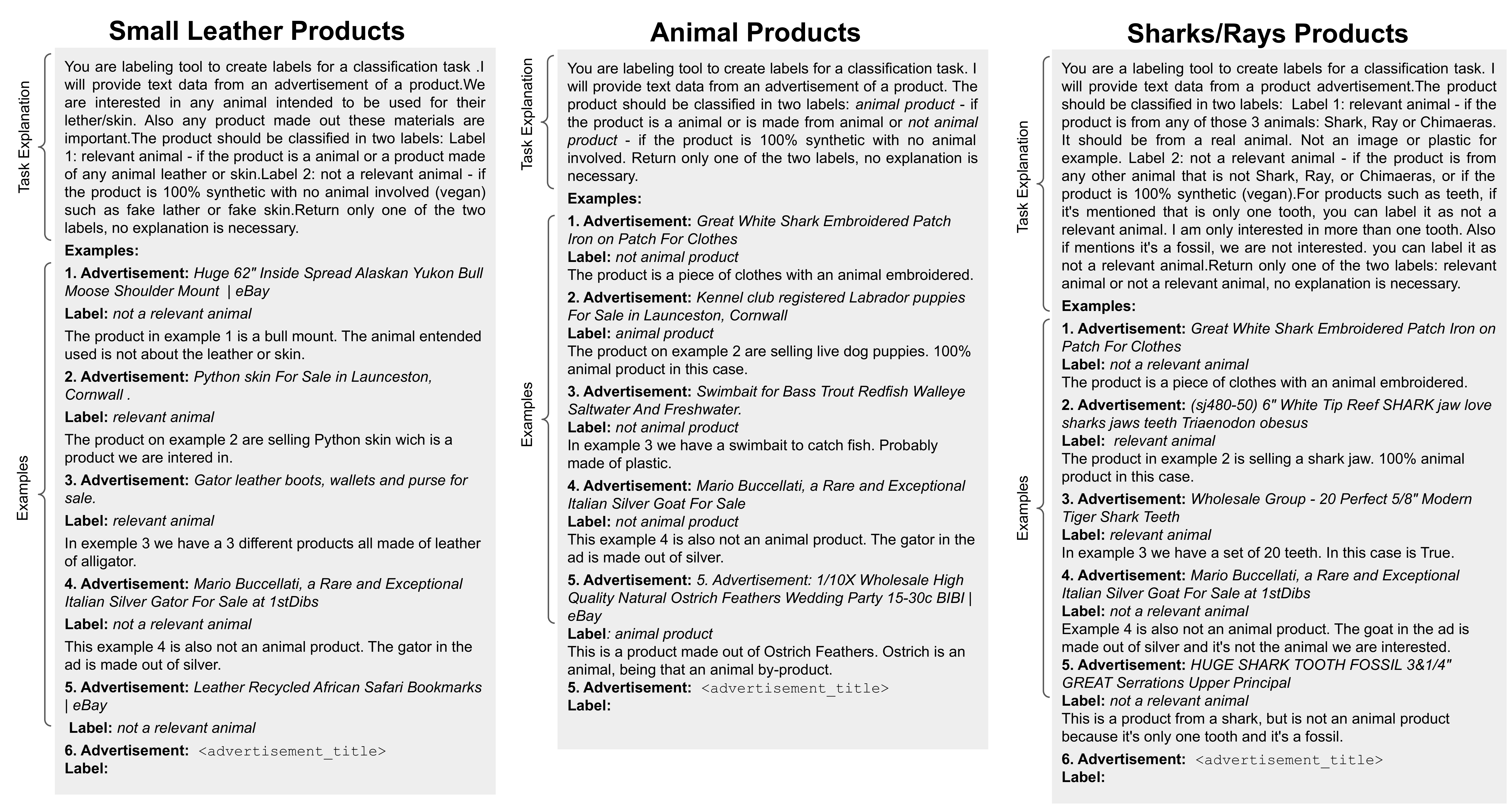}
    \vspace{-.7cm}
        \caption{\review{Few-shot prompts used to generate labels using GPT-4 for the three research questions. Exemplary instances with labels and explanations are included to demonstrate the task to the LLM.}}
        \Description{}
            \vspace{-.3cm}
        \label{fig:Prompt}
\end{figure*}

Figure~\ref{fig:Prompt} shows the prompts used to label data for three research questions. They   include a brief explanation of the task, followed by examples of ad titles with their corresponding labels and a brief rationale. \review{For example, for sharks and rays, the prompt explains which animals should be classified as "relevant" and states that fossils are not relevant for the task. We include positive and negative examples to demonstrate how the task should be carried out--demonstration is a prompt engineering strategy shown to be effective for complex tasks~\cite{white2023prompt,llmWrangle2022}.} Note that prompts can be easily customized by domain experts to match the selection criteria for a research question.

\myparagraph{Selecting Representative Samples from an Ad Collection} 
For wildlife ads, selecting a representative set of ads to label is difficult due to the stark imbalance in the data: there are very few relevant ads. In some cases, as few as 5\% of the ads retrieved by crawlers include animals or animal parts \cite{barbosa2024flexiblescalableapproachcollecting}.
This imbalance arises from the need to maximize recall during the data collection and avoid missing relevant data. 
Therefore, while constructing a sample, we must avoid selecting too many ads in the negative class while ensuring that a sufficient number of representatives of the rare class are selected. It is also important that both positive and negative samples are diverse. For instance, various shark parts can be sold, such as teeth, jaws, or full taxidermy. Similarly, when considering negative samples, we must recognize that items like shark plush toys are irrelevant, but we also need to account for other options such as shark-themed stamps, books, and countless other possibilities.

\myparagraph{Problem Definition}    
Let $D$ be an ad collection containing $N$ items, where $N_{\text{pos}}$ represents the number of relevant ads (positive examples) and $N_{\text{neg}}$ represents the number of irrelevant ads (negative examples).
We want to select a subset $S \subseteq D$ such that $S$ contains $m$ samples that include a sufficient number of positive examples.
Specifically, let $N_{pos}/N_{neg}=k$, $S_{\text{pos}}$ represent the number of positive samples in $S$, and $S_{\text{neg}}$ represent the number of negative samples in $S$.
We aim to construct a sample in which the ratio $S_{pos}/S_{neg} \gg k$.
Furthermore, since our objective is to maximize the accuracy of a machine-learning model trained using the selected sample $S$, $S$ should be diverse and representative of the original dataset $D$. 

In Section~\ref{subsec:sampling}, we discuss widely-used approaches for sampling and their limitations. We then introduce our approach, \ourapproach, in Section~\ref{subsec:proposed}.

\subsection{Sampling Ads to be Labeled}
\label{subsec:sampling}

\myparagraph{Random Sampling}
Random sampling is a straightforward strategy to select items from 
a large collection.
It is simple to implement and helps prevent bias towards any particular class since items are selected at random.
However, since the sample maintains the distribution of the original data,  
in datasets where one class heavily outweighs others in terms of frequency, random sampling may result in a subset that predominantly consists of instances from the majority class, thus failing to adequately represent the minority classes. This, in turn, can lead to ineffective classifiers. For example, as shown in Table~\ref{tab:aggregated_metrics}, classifiers trained using random samples have the lowest scores (F1-score of 0.465).
 
\myparagraph{Knowledge-Based Sampling}
Another possible approach to sampling data that increases the chances of selecting underrepresented data points is to exploit prior knowledge about the dataset. This can be achieved by identifying 
%categories or 
characteristics of data points that are less frequently represented (i.e., in our scenario, positive examples) and strategically sample from those areas. 
For example, we can search for ads that contain certain keywords that are strong indicators of an animal-related product, such as searching for a body part that is probably related to an animal such as a \textit{barb} or \textit{rostrum} to identify Rays.
Even though this approach can help increase the number of positive examples in the sample, diversity may suffer as the samples will be biased toward the keywords used. This, in turn, may lead to overfitting during model training. The limitations of this approach can be seen in Table~\ref{tab:aggregated_metrics}: the recall for models trained with this sampling approach is low, suggesting that the model misclassifies a significant number of positive instances.   

\subsection{Proposed Approach: Clustering-Based Pseudo-Label Generation}
\label{subsec:proposed}

We propose \ourapproach, a new approach to select a set of representative samples from an unbalanced collection. Our goal is to select a sufficient number of diverse samples that can be used to train an effective classifier.
Figure \ref{fig:csb-data-generator} provides an overview of \ourapproach that takes as input a set of e-commerce ads that are retrieved by a web crawler (Section~\ref{sec:approach-overview}).

\paragraph{Clustering} To ensure that a diverse set of ads is selected, the ads are first clustered. Different clustering techniques can be used. For our current implementation, use used a topic modeling method \cite{vayansky2020review}. 

\paragraph{Cluster Selection}  Given the set of clusters, the next step is to select the clusters from which to draw samples. 
In this step, LLMs can be useful as zero-shot classifiers to automatically determine the label for a sample obtained from a given cluster (Figure~\ref{fig:Prompt}). 
Here we face a challenge: how to determine which clusters and samples to select. 
Recall that our goal is to obtain a diverse set of both positive and negative examples. So, to maximize this diversity and create a balanced sample while minimizing the number of required LLMs inferences, we propose a method that learns to select \emph{good} samples that balances exploration and exploitation of the clusters, and through active learning, builds and incrementally refines a specialized classifier

As detailed in Algorithm \ref{algorithm:thompson}, 
we employ Thompson sampling for cluster selection.
While there are other multi-armed bandit (MAB) approaches, such as Upper Confidence Bound (UCB),  empirical studies have shown that Thompson sampling serves as a robust baseline for MAB problems \cite{chapelle2011empirical}. 
Algorithm \ref{algorithm:thompson} starts by initializing the variables and parameters (lines 1-3). Next, it samples a cluster from the beta distribution (lines 5-8). In line 9, the reward is computed based on the improvement of the model trained using the data from the cluster selected (further details below). Based on the reward obtained, it updates the number of wins or losses (lines 10-14).
Finally, it applies the decay factor to the wins and losses arrays after updating them based on the observed reward (lines 15-16), reducing the influence of past observations on the current state of the algorithm.

\begin{algorithm}[t]
\caption{Cluster Selection}
\small
\begin{algorithmic}[1]
\State Initialize an empty array to store the number of wins: $\text{wins}[i] = 0$ for $i = 1$ to $K$.
\State Initialize an empty array to store the number of losses: $\text{losses}[i] = 0$ for $i = 1$ to $K$.
\State Set decay factor $\delta \in (0, 1)$ (e.g., $\delta = 0.99$).
\For{$t = 1$ to $T$}
    \For{$i = 1$ to $K$}
        \State Sample $\theta_i(t)$ from Beta distribution with parameters $\alpha + \text{wins}[i]$, $\beta + \text{losses}[i]$.
    \EndFor
    \State Select cluster $a_t = \arg \max_i \theta_i(t)$. 
    \State Observe reward $r_t$.
    \If{$r_t > 0$}
        \State Increment wins: $\text{wins}[a_t] \leftarrow \text{wins}[a_t] + 1$. 
    \Else
        \State Increment losses: $\text{losses}[a_t] \leftarrow \text{losses}[a_t] + 1$.
    \EndIf
    \State Apply decay: $\text{wins}[i] \leftarrow \text{wins}[i] \times \delta$ for $i = 1$ to $K$.
    \State Apply decay: $\text{losses}[i] \leftarrow \text{losses}[i] \times \delta$ for $i = 1$ to $K$.
\EndFor
\end{algorithmic}
\label{algorithm:thompson}
\end{algorithm}

\paragraph{Sample Selection} After a cluster is selected, $N$ samples are drawn from that cluster. In the first iteration, these samples are selected randomly. In subsequent iterations, the samples are selected using a fine-tuned model trained using the samples selected (and labeled) in prior iterations (see \textit{Retrain Model} below).
To sample $N$ data points, we prioritize selecting points from positively labeled data up to a maximum number (given as a parameter). If there are not enough positive samples available to reach the maximum parameter, we take as many samples as possible and then fill the remaining slots with samples from the negatively labeled data.
This is performed to increase the chances of selecting positive data and balance the training data.

\paragraph{LLM Pseudo-Labeling} After selecting the N samples from the cluster, we obtain the label from an LLM through in-context learning (few-shot) as illustrated in Figure~\ref{fig:Prompt}.

\paragraph{Retrain Model} The labeled data is then used to fine-tune a base model. After training, we evaluate the model's performance against the validation gold data. The results are used to decide how to update the Thompson sampling parameters, specifically the counts of \textit{wins} and \textit{losses}.
There are two scenarios:\\
\textit{(1)} If the model outperforms the baseline score, we set the Thompson sampling (TS) reward $r_t$ to 1 (line 9), which increments the \textit{wins} for the selected cluster (line 11). This model is designated as the new base model for the next fine-tuning. In the first iteration, when there is no existing model, a metric and a baseline score parameter are used for comparison to generate the reward.\\
\textit{(2)} If the model performs worse than the baseline, we increment the losses for the selected cluster by setting the TS reward $r_t$ to 0 (line~9) and revert to the previous base model for further fine-tuning.

After retraining the model, we start a new iteration and repeat all previous steps, except the initial clustering, until the fine-tuned model achieves the required performance or the budget is reached (e.g., the number of iterations or LLM inferences). The final output is a diverse set of pseudo-labeled ads to be used to train a more robust machine-learning model.

%% file: approach-overview.tex
\begin{figure}[t]
\centering
\includegraphics[width=0.95\columnwidth]{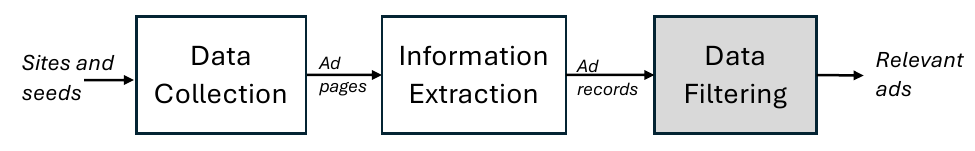}
\caption{Wildlife Data Curation Pipeline}
\label{fig:data-collection-pipeline}
\end{figure}

The data curation pipeline employed to obtain relevant wildlife ads is depicted in \autoref{fig:data-collection-pipeline}. This pipeline is designed to gather relevant product pages from a pre-defined set of online marketplaces. 

\myparagraph{Data Collection} 
To collect ads from marketplaces, we use the ACHE focused web crawler~\cite{chang2022survey, ache@github}. 
The crawler is given as input a set of \textit{seed} URLs that serve as entry points for the crawl, e.g., the sites for the marketplaces, as well as a list of species of interest. 
To retrieve the ads, the crawler can navigate the sites or leverage their search interfaces to obtain ad pages directly. For example, to obtain ads related to \emph{sharks} on \texttt{ebay.com}, the crawler can issue the following HTTP request:
{\color{blue}\url{/sch/i.html_from=R40&_nkw=}\color{red}\textbf{\texttt{shark}}\color{blue}\url{&_sacat}}.
An effective strategy to collect ads is to derive seed links that cover different species and variations of the terms used for those species (e.g., for sharks, we can also search for \textit{dogfish}, which is type of shark).
The crawler downloads pages from the seed URLs and recursively follows links extracted from these pages.

\myparagraph{Information Extraction} 
Given the pages collected by the crawler, the Information Extraction component identifies within these pages information associated with the product so that we can create, for each product, a record containing its relevant attributes, e.g., \textit{Price, Seller Name, Product Type, Description, and image}. Different tools and methods can be used for extraction~\cite{beautifulsoup,llm-ensemble@sigir2024,mave@wsdm2022}.

\myparagraph{Data Filtering}
The final step of the pipeline is to select the relevant ads that match the selection criteria for the research question, and prune ads that do not include animals or that are not derived from living animals. As discussed in Section~\ref{sec:intro}, machine learning classifiers are a natural choice for this task, but obtaining labeled data to create these classifiers is costly and time-consuming. 
%existing approaches to data filtering are not scalable and have hampered the study of wildlife trafficking. 
In what follows, we present our approach to this challenge.
%In this paper, we address the challenge of accurately distinguishing animal products from a range of unrelated items, such as postcards, plush toys, and others, which are frequently retrieved alongside wildlife products. Our approach focuses on pseudo-labeling training data to construct a good machine-learning model to ensure reliable data for domain experts. We provide an explanation of our methodology for creating such data and training a model tailored to accurately detect and classify advertisements featuring animal products.

%% file: experimental-evaluation.tex
We evaluated our framework's  effectiveness and robustness in deriving classifiers through three (real) research questions in the wildlife domain.
Leather products (LP) and shark products (SP), introduced in Section~\ref{sec:intro} and detailed in Section \ref{sec:use-cases}, consist of specific questions that focus on a small set of species and products. Animal products (AP) encompasses a more general 
classification task to determine whether an ad contains an animal-derived product, consisting of a large collection covering many species that is useful to assess the scalability of \ourapproach.

\subsection{Experimental Setup}

\myparagraph{Data Collection and Validation Data}
We use the open-source ACHE crawler to perform a \textit{scoped crawl}
\cite{ache@github}, i.e.,  given a seed list of Web pages, the crawler recursively follows all the links in these pages that belong to the given domain. 
The datasets used in the evaluation are summarized in Table~\ref{tab:datasets} and described below.

\myparagraphemph{Shark products (SP)} 
We crawled the ads pages on eBay.com, using initial seeds from a list of 3 animals commonly traded as trophies: sharks, rays, and chimeras, with their respective common names and paired with body parts related to each animal. 
A total of 14,533 valid ads were collected. Invalid ads, e.g., ads with no price, are excluded from the collection. 

\myparagraphemph{Leather products (LP)} 
To create the seed list for leather products, we used the names of 48 animals listed in the seized wildlife and their intended uses dataset~\cite{stringham2021dataset}, and obtained ads from eBay.
The crawler collected 152,495 valid ads. 

\begin{table}[t]
\small
   \caption{\review{Research questions, evaluation datasets, and the respective number of ads.}}
\label{tab:datasets}
\review{
\begin{tabular}{p{.45\columnwidth} p{.25\columnwidth} p{.16\columnwidth}}
\toprule
\centering
\multirow{1}{*}{\textbf{Research Question}} & \multirow{1}{*}{\textbf{Dataset}} & \textbf{Number of Ads} \\ %\hline
\midrule
Identify products derived from sharks, rays and chimaeras & Shark products (SP) & 14,533\\
\midrule
Identify leather products derived from endangered animals & Leather products (LP) & 152,495 \\ 
\midrule
Identify animal-derived products & Animal products (AP) & 699,907\\ \bottomrule
\end{tabular}
}
\end{table}

\myparagraphemph{\review{Animal products (AP)}}
The animal products ad collection consists of 699,907 valid ads, gathered over a one month period from 33 different e-commerce sites, with the goal of covering  
mammals, birds, and reptiles, including 263 different endangered species listed on CITES~\cite{citesreport2022}.
This data was collected to help the experts study and quantify the trade for different species.  In contrast to the more specific questions for the small leather and shark ads, 
this task aims to determine whether an ad is for an animal part or animal-derived product.  Besides demonstrating generalizability, we also use this larger dataset to study the scalability of our approach.

To evaluate the derived models, we used approximately 500 ads manually labeled by domain experts for each research question.

\input{performance_table}

\myparagraph{Baselines} To assess the effectiveness of our approach, we compare it against the sampling strategies described
in Section~\ref{subsec:sampling} and an \review{active learning algorithm}. For \textit{Random Sampling (RS)}, we draw a uniform random sample of ads from the complete dataset generated from each use case. We configured \textit{Knowledge-based Sampling (KBS)} as follows.
For \emph{LP}, we used string matching to find the possible animal names present in the initial set of seeds provided to the
crawler and some keywords related to the products known to be derived from such animals. The keywords contain species scientific names, common names, their uses, and products. This data was
sourced from a larger dataset on seized wildlife and their intended
uses \cite{stringham2021dataset}. The domain experts provided us with keywords related
to possible products that are usually found for sale based on each of
the animals in the list. The final list contained around 226 product-type keywords and 250 animal names. With this list, we search for
3 options for string matching. 1. animal species names 2. animal
common names, and 3. products. The ad should always contain an animal name (\textit{species name} or \textit{common name}) and a \textit{product name} to be selected as a positive example. After selecting the candidate positive ad, we randomly select the samples from the
collection. 

We used a similar approach for \emph{SP}. 
The domain experts provided us with keywords related to possible animal parts used for such trades. The final list contained around 11 animal parts and 19 names related to each of the 3 species. With this list, we search for: 1) animal names and 2) animal body parts. If the ad title contains the
animal name and  body part,  the ad is selected as a
positive candidate. 
For \emph{AP}, the same approach is used for partial and complete matches of animal species and common names.
The list of positive candidates is used to sample the data. 

The \emph{active learning (ACTL)} algorithm we used the small-text library \cite{schroder2023small} configured for a binary text classification task using \emph{bert-base-uncased} as the base model. It utilizes a pool-based active learner with a greedy coreset query strategy, which selectively queries the most informative data points for labeling, optimizing the learning process with fewer labeled examples. The model is trained leveraging GPU for efficient processing, and is designed to handle class imbalance by applying balanced class weights.

\myparagraph{\ourapproach Setup} \ourapproach uses \textit{bert-base-uncased} as the pre-trained model to be fine-tuned during the active learning iterations for shark and leather products. \review{For animal products, since some of the ads are extracted from sites that use languages other than English, \ourapproach uses \textit{bert-base-multilingual-cased}}. 
We use F1-score as the evaluation metric and the baseline is set initially to 0.5; after the first model is trained the baseline is updated with the performance of the F-1 score (see details in Section~\ref{subsec:proposed}).
All labels are generated by GPT-4 through in-context learning (few-shot), as described in Section~\ref{sec:training-data-generation}.

\input{sigmod_submited/models}

\subsection{Sampling Strategies and Active Learning}

We have assessed the effectiveness of different sampling strategies as well as of a state-of-the-art active learning approach (ACTL) \cite{schroder2023small, sener2017active}, using Precision, Recall, F1-Score, and Accuracy. The results are summarized in Table~\ref{tab:aggregated_metrics}.
The models trained with \ourapproach-labeled data consistently outperform by a large margin the other sampling strategies (KBS and RS) in terms of F-measure: they attain F-measures between .85 and .94 for small leather products,  between .87 and .89 for shark products, and between .76 and .81 for animal products.

Not surprisingly, the models trained using randomly sampled data performed poorly overall due to overfitting the unbalanced class present in the data. \review{The multi-modal models trained with the random sample obtained F1-score close to zero for the three research questions, but performed reasonably well (F1 score between .74 and .77) for the text-based models for animal and shark products. Also note that except for leather products, the performance of the multi-modal models is worse than that of the text-based counterparts. This suggests that images do not always help improve the classification. For each of the ads only one image is saved (if available) and we cannot be certain whether 
the image posted by the seller is accurate and reflects what they are selling. Therefore, the choice of multi-modal and text-based models must be informed by the research question and the characteristics of the ads for relevant products.}

The results show that while the KBS multi-modal model attains high precision for sharks products, it does so at the cost of recall. This can be explained by the bias introduced in the user-provided keywords towards a subset of the positive examples. For the animal product models, the complexity due to the large number of animals and insufficient prior knowledge about all animals increased the bias, causing the KBS models to perform worse than RS models.

\ourapproach outperformed ACTL for leather and animal products, both for the text-based and multi-modal models. For shark products, ACTL performed slightly better for the text-based model and attained performance similar to \ourapproach for the multi-modal model. However, we encountered scalability issues with ACTL. For animal products, which includes a larger number of ads (699,907), ACTL was not able to complete the learning process after 72 hours, generating only 3,000 out of the 4000 samples generated by the other baselines. Like \ourapproach, the coreset-based active learning of ACTL aims to generate a representative sample of the data, but it is more computationally expensive: 
it generates embeddings for all data points, and calculates pairwise distances (or similarities) between points in the embedding space--scaling quadratically with dataset size. This, combined with the need to repeatedly retrain the model in each iteration, limits the scalability of the approach.
\ourapproach is able to generate representative samples at a substantially lower cost: it performs clustering only once, at the beginning of the process, and performs inference only over the selected cluster for each iteration instead of over complete dataset. We can see the differences between the two approaches in terms of F1-score and processing time in Table~\ref{tab:processing_time}. These numbers show that \ourapproach provides a good cost-benefit trade-off, but also suggest that there is room for improvement. 
In future work, we plan to explore if alternative embedding-based clustering strategies 
can improve the target classification models.

We have also analyzed the precision and recall for the different classes -- positive and negative -- for the baselines and \ourapproach. Table~\ref{tab:metrics_sharks} shows the numbers for \ourapproach, KBS, RS, and ACTL for animal products using the text-based model. For all baselines, the precision and recall for the negative class are higher than for the positive, as expected. While KBS, RS and ACTL have lower recall and precision than our approach for both classes, the difference is larger for the the positive class:  
\ourapproach attains 0.79 for recall and 0.84 for precision.
The fact that \ourapproach attains a balance of precision and recall for both positive and negative classes underscores its
effectiveness at selecting the training samples. The models trained with the \ourapproach samples can identify most of the relevant ads without making excessive false positive errors.
We have observed a similar pattern for the classification tasks for shark and leather products.   

Overall, our experiments show that, unlike the other sampling strategies, \ourapproach maintains a balance between precision and recall, suggesting that the samples it selects are diverse and representative of the whole data. 

\begin{table}[t]
\small
   \centering
   \caption{\review{Comparison of \ourapproach and Active Learning baseline regarding processing time (in hours) versus F1 score for various dataset sizes. The best performance, characterized by the highest F1 score and lowest processing time, is highlighted.}}
\label{tab:processing_time}
\begin{tabular}{ccccc}
\toprule
\textbf{Dataset} & \textbf{Size} & \textbf{Approach} & \textbf{Time (hours)} & \textbf{F1 score} \\
\midrule
\centering
\multirow{2}{*}{Animal} & \multirow{2}{*}{700k}  & ACTL & 72* & 0.48 \\
                     &                         & LTS  &  \cellcolor{green!10} \textbf{4.81} & \cellcolor{green!10} \textbf{0.82} \\
\midrule
\centering
\multirow{2}{*}{Leather} & \multirow{2}{*}{152k} & ACTL & 8.99 & 0.82 \\
                     &                         & LTS  & \cellcolor{orange!25} \textbf{1.15} & \cellcolor{orange!25} \textbf{0.86} \\
\midrule
\centering
\multirow{2}{*}{Sharks} & \multirow{2}{*}{15k} & ACTL & 1.29 & \cellcolor{blue!15} \textbf{0.94} \\
                    &                        & LTS  &  \cellcolor{blue!15}\textbf{0.77} &  0.87 \\
\bottomrule
\end{tabular}
\end{table}

\begin{table*}[t]
\centering
\caption{Comparison of the precision and recall of text-based models derived by different baselines for animal products for classifying positive and negative ads.}
\label{tab:metrics_sharks}
\begin{small}
\centering
\review{
\begin{tabular}{lcccccccc}
    \toprule
     & \textbf{Precision} & \textbf{Recall} & \textbf{F1 Score} & \textbf{Precision} & \textbf{Recall} & \textbf{F1 Score} & \multirow{2}{*}{\textbf{Accuracy}} \\
     & (Class 0) & (Class 0) & (Class 0) & (Class 1) & (Class 1) & (Class 1) & \\
     \midrule
    \textbf{LTS} & \cellcolor{green!10}{0.948} & \cellcolor{green!10}{0.962} & \cellcolor{green!10}0.955 & \cellcolor{green!10}0.842 & \cellcolor{green!10}{0.792} & \cellcolor{green!10}{0.816} & \cellcolor{green!10}0.928 \\
    \textbf{KBS} & 0.906 & 0.955 & 0.930 & 0.775 & 0.614 & 0.685 & 0.886 \\
    \textbf{RS} & 0.937 & 0.932 & 0.934 & 0.738 & 0.752 & 0.745 & 0.896 \\
    \textbf{ACTL} & 0.898 & 0.960 & 0.928 & 0.784 & 0.574 & 0.663 & 0.882 \\
\bottomrule
\end{tabular}
}
\end{small}
\end{table*}

\subsection{Specialized versus Foundation Models} 

Table~\ref{table:llms_baseline_combined} shows a performance comparison between the text-based model created with data sampled using \ourapproach (\ourapproach-text) against state-of-the-art language models in the zero-shot setting--LLAMA3, GPT-4, and LLAVA-MM. 
For small leather products, \ourapproach-text achieved an F1-Score of 0.940, outperforming LLAMA3 (0.781), GPT-4 (0.850), and LLAVA-MM (0.472).
For shark trade, \ourapproach-text achieved an F1-Score of 0.89, outperforming the open-source models and approaching the performance of GPT-4 (0.946).
The superior performance of GPT-4 in the shark products task is not surprising as it is trained on a vast and diverse dataset and encompasses 1.76 trillion parameters. 

\paragraph{Number of Parameters} The efficiency of the \ourapproach-text model is notable: it utilizes only 110 million parameters.
This is in contrast to  GPT-4 which employs 1.76 trillion parameters.
This makes \ourapproach suitable for scenarios where computational resources are limited, enabling its use in applications without the overhead (and cost) required by larger models. Despite its lower parameter count, \ourapproach-text achieves a high F1-Score for the three tasks, confirming the effectiveness of the \ourapproach sampling strategy.

\paragraph{Inference Cost}
To create the samples for shark and small leather products, we need only 2,000 GPT-4 API requests for each question, a total of \$40 US dollars (around 2 cents per request). \review{For animal products, we issue 4,000 GPT-4 requests--totaling \$40 US dollars}. In contrast, to classify the ads for small leather products using GPT-4 it costs roughly \$3,350, for shark trade, the total cost is around \$320 and for \review{animal products \$14,000}. For sampling, \ourapproach incurs a one-time cost to create classifiers that can be subsequently used for multiple inferences; in contrast, we need to pay to classify each individual ad using GPT-4.

Overall, the model produced using \ourapproach performs substantially better than LLAMA and LLAVA-MM. It  performs comparably to GPT4 while using only 6.2\% of the total number of parameters and at a small fraction of the cost. 
This suggests that \ourapproach is a cost-effective approach for the creation of classifiers to be used in resource-constrained scenarios such as the study of wildlife trafficking.

\begin{table*}[t]
    \centering
    \caption{\review{Performance comparison (using F-1 measure) of the text-based model trained using \ourapproach (\ourapproach-text) against zero-shot classification performed by different LLMs. The best performance for each task is highlighted.}} \label{table:llms_baseline_combined}
        \centering
        \begin{tabular}{ccccc}
            \toprule
            \textbf{Task} & \textbf{LLAMA3 (8 Bi)} & \textbf{GPT4 (1.76 Tri)}  & \textbf{LLAVA-MM (7 Bi)} & \textbf{\ourapproach-text (110 Mi)} \\
            \midrule
            \review{Animal Product}  & \review{0.556} &  \cellcolor{green!10} \review{0.827} &  \review{0.481} & \review{0.816} \\
            Leather Products & 0.781 & 0.850 & 0.472 & \cellcolor{orange!25} 0.855 \\
            Shark Products & 0.659 & \cellcolor{blue!15}0.946 & 0.247 &  0.873 \\    
            \bottomrule
        \end{tabular}
\end{table*}

\begin{table}[t]
    \centering
    \caption{\review{Ablation of LLM labelers used to generate training data for \ourapproach. The table shows, for each dataset, the F1-score of the text-based model trained with the generated training data. The highest F1 score for each task is highlighted.}}
        \centering
        \begin{small}
        \review{
        \begin{tabular}{cccccc}
            \toprule 
            \multirow{2}{*}{\textbf{Task}} & \textbf{LLAMA} & \textbf{LLAMA} &  \textbf{LLAMA} & \textbf{GPT} & \textbf{GPT} \\
            {} & \textbf{3.1:8B} & \textbf{3.1:70B} & \textbf{3.3:70B} & \textbf{4o-mini} & \textbf{4}\\
            \midrule
            Animal Product & 0.4 & 0.474 & 0.54 & 0.59 & \cellcolor{green!10} 0.816  \\
            Small Leather & 0.616 & 0.619 & 0.64 & 0.691 & \cellcolor{orange!25} 0.855 \\
            Sharks & 0.591 & 0.63 & 0.600 & 0.6846 &  \cellcolor{blue!15} 0.873 \\
            \bottomrule
        \end{tabular}
        }
        \end{small}
    \label{table:ablation}
\end{table}

\vspace{-.2cm}
\subsection{Ablation using Different LLMs to Produce Pseudo-Labels}

To assess the impact of model size for deriving the pseudo-labels, we evaluate the performance of the text-based model created by \ourapproach using five different LLMs: LLAMA3.1 8B, LLAMA3.1 70B, LLAMA3.3 70B, GPT4o-mini, and GPT4.  
As shown in Table \ref{table:ablation}, the F1 score of the models 
for all three tasks increases as the size of the LLM increases. Not surprisingly, GPT4 leads
to the highest overall F1 (greater than 0.81).
The smaller models perform substantially worse, attaining F1 measures between 0.4 and 0.69. These results underscore the importance of using larger models to produce pseudo-labels, as high-quality labels are essential for the learning process.

\begin{table*}[t]
\centering
\caption{\review{Runtime breakdown for the different components of \ourapproach. Sample size per iteration is fixed and clustering is performed just once during the sample generation process. We report the mean and standard deviation of the time required for inference, labeling and training over 10 iterations.}}
\label{tab:time}
\resizebox{\textwidth}{!}{
\begin{tabular}{ccc|cccc|c}
\toprule
\multirow{2}{*}{\textbf{Dataset}} & \multirow{2}{*}{\textbf{Clustering}}    & \multirow{2}{*}{\textbf{\parbox{2cm}{\centering Sample\\Size}}}       & \multicolumn{4}{c}{\textbf{Running 10 iterations}} & \multirow{2}{*}{\textbf{Total}} \\
                         &                      &                         & \textbf{Selected Cluster size}   & \textbf{Inference Time (s)}    & \textbf{Labeling Time (s)}       & \textbf{Training Time (s)}      &                        \\
\midrule
AP - 700k & 495 (s) & 200 & 106,625.8 ± 34057.8 &954.334 ± 310.857 & 84.482 ± 8.56 &89.541 ± 16.45 & 180 Min \\
\hline
SP- 15k & 19.62 (s) & 200 & 3773 ± 869.65 & 33.62888 ± 7.911 & 96.725 ± 27.68  &85.209 ± 15.105 & 35.694 Min \\
\bottomrule
\end{tabular}
}
\end{table*}

\vspace{-.2cm}
\subsection{Execution Time Breakdown}

To study the computational costs and scalability of \ourapproach, we contrast the costs for a small and a large task: shark products (14,533 ads) and animal products (699,907 ads). We measured the time needed to run the \ourapproach  components for deriving the same number (2,000) of labeled ads for both tasks. To perform a fair comparison, we also used the same base model for the active learning process, \emph{bert-based-uncased}. The results are summarized in Table \ref{tab:time}.%
\footnote{The running time in Table~\ref{tab:time} is lower than the value reported in Table~\ref{tab:processing_time}, since for the former \ourapproach is configured to derive half of the labeled ads and uses a smaller base model.}  

As expected, clustering takes longer for the larger data, however this step is executed only once at the start of the process and it takes less than 4\% of the total time. The main bottleneck is inference, required to label ads in the clusters. The inference cost depends on the size of the data and the complexity of the based model used.
We also measured the cost of pseudo-labeling (using gpt4o-mini) and training a model with the labeled data: since the sample size is fixed, the costs are similar for both datasets. While the costs (in time and money) of running \ourapproach are adequate for our application, we plan to explore ways to optimize the pipeline and, in particular, to reduce the inference time.

%% file: performance_table.tex
\begin{table*}[t]
\small
\centering
\caption{\review{The performance of text-based and multi-modal models created using data generated by various baseline strategies for three classification tasks: animal products, leather products, and shark products. The best performance for each metric is highlighted in different colors (one color per task).}}
% \jf{should the 0.842 precision for LTS be highlighed?}}
\resizebox{\textwidth}{!}{
\begin{tabular}{llcccccccccccc}
\toprule
    & & \multicolumn{4}{|c}{\textbf{\review{Animal Products}}} & \multicolumn{4}{|c}{\textbf{Leather Products}} &  \multicolumn{4}{|c}{\textbf{Sharks Products}} \\ 
    \textbf{Model} & \textbf{Metric} & \review{\textbf{RS}} & \review{\textbf{KBS}} & \review{\textbf{ACTL}} & \review{\textbf{\ourapproach}} & \textbf{RS} & \textbf{KBS} & \review{\textbf{ACTL}} & \textbf{\ourapproach}  & \textbf{RS} & \textbf{KBS} & \review{\textbf{ACTL}} & \textbf{\ourapproach}\\
    % \hline
    \midrule
    \centering
    \multirow{4}{*}{\textbf{Text-Based}} & \multicolumn{1}{l}{Accuracy}  & \multicolumn{1}{c}{0.8957} & \multicolumn{1}{c}{0.8857} & \multicolumn{1}{c}{0.882*} & \cellcolor{green!10}0.928 & \multicolumn{1}{c}{0.588} & \multicolumn{1}{c}{0.790} & 0.82 & \multicolumn{1}{c}{\cellcolor{orange!25}0.854} & \multicolumn{1}{c}{0.915} &  \multicolumn{1}{c}{0.923} & \cellcolor{blue!15}0.969 & 0.946 \\ 

             & \multicolumn{1}{c}{F1-Score} & \multicolumn{1}{c}{0.745} & \multicolumn{1}{c}{0.685} & \multicolumn{1}{c}{0.663*} & \cellcolor{green!10}0.816 & \multicolumn{1}{c}{0.465} & \multicolumn{1}{c}{0.776} & 0.822 & \multicolumn{1}{c}{\cellcolor{orange!25}0.855} & \multicolumn{1}{c}{0.774} &  \multicolumn{1}{c}{0.780} & \cellcolor{blue!15}0.924 & 0.872 \\
             
             & \multicolumn{1}{c}{Precision} & \multicolumn{1}{c}{0.737} & \multicolumn{1}{c}{0.775} & \multicolumn{1}{c}{ 0.784*} & \cellcolor{green!10}0.842 & \multicolumn{1}{c}{0.708} & \multicolumn{1}{c}{0.833} & 0.722 & \multicolumn{1}{c}{\cellcolor{orange!25}0.858} & \multicolumn{1}{c}{0.857} &  \multicolumn{1}{c}{0.944} & \cellcolor{blue!15}0.958 & 0.844 \\
             
             & \multicolumn{1}{c}{Recall} & \multicolumn{1}{c}{0.752} & \multicolumn{1}{c}{0.614} & \multicolumn{1}{c}{0.574*}& \cellcolor{green!10}0.792 & \multicolumn{1}{c}{0.588} & \multicolumn{1}{c}{0.790} & 0.954 & \multicolumn{1}{c}{\cellcolor{orange!25}0.854} & \multicolumn{1}{c}{0.705} &  \multicolumn{1}{c}{0.666} & 0.892 & \cellcolor{blue!15}0.902 \\ \hline
             
    \multirow{4}{*}{\textbf{Text-Image}} & \multicolumn{1}{l}{Accuracy} & \multicolumn{1}{c}{0.174} & \multicolumn{1}{c}{0.140} & \multicolumn{1}{c}{0.83*} &\cellcolor{green!10}0.921& \multicolumn{1}{c}{0.568} & \multicolumn{1}{c}{0.838} & 0.872 & \multicolumn{1}{c}{\cellcolor{orange!25}0.946} & \multicolumn{1}{c}{0.7955} &  \multicolumn{1}{c}{0.926} &0.956 & \multicolumn{1}{c}{\cellcolor{blue!15}0.956} \\

             & \multicolumn{1}{c}{F1-Score} & \multicolumn{1}{c}{0.297} & \multicolumn{1}{c}{0.216} & \multicolumn{1}{c}{0.596*} & \cellcolor{green!10}0.765& \multicolumn{1}{c}{0.027} & \multicolumn{1}{c}{0.780} & 0.851& \multicolumn{1}{c}{\cellcolor{orange!25}0.940} &  \multicolumn{1}{c}{0.0} &  \multicolumn{1}{c}{0.789} & \cellcolor{blue!15}0.893 & \multicolumn{1}{c}{0.89} \\
              
             & \multicolumn{1}{c}{Precision} & \multicolumn{1}{c}{0.175} & \multicolumn{1}{c}{0.128} & \multicolumn{1}{c}{0.51*} & \cellcolor{green!10}0.8 & \multicolumn{1}{c}{0.750} & \multicolumn{1}{c}{\cellcolor{orange!25}0.945} &0.863 &  \multicolumn{1}{c}{0.913} & \multicolumn{1}{c}{0.0} &  \multicolumn{1}{c}{\cellcolor{blue!15}0.945} &0.901 & \multicolumn{1}{c}{0.908} \\
              
             & \multicolumn{1}{c}{Recall} & \multicolumn{1}{c}{1} & \multicolumn{1}{c}{0.676} & \multicolumn{1}{c}{0.718*} & \cellcolor{green!10}0.732 & \multicolumn{1}{c}{0.014} & \multicolumn{1}{c}{0.661} & 0.839 & \multicolumn{1}{c}{\cellcolor{orange!25}0.968} & \multicolumn{1}{c}{0.0} &  \multicolumn{1}{c}{0.676} & \cellcolor{blue!15}0.884 & \multicolumn{1}{c}{0.873} \\
    \hline
\end{tabular}
}
\label{tab:aggregated_metrics}
% \caption{Performance Metrics of Text-Based and Text-Image Models Across Different Sampling Methods in Detecting Animal Advertisements}
% \vspace{0.3em} % Adjusts the space between the table and the note

{\footnotesize 
    \review{* Timeout - After 78 hours only 3,000 training examples were generated and used to train the models.}}
\end{table*}

%'Precision: 0.9578947368421052, Recall: 0.8921568627450981, F1 Score: 0.9238578680203046, Accuracy Score: 0.969939879759519'

%% file: sigmod_submited/models.tex
\begin{figure}[t]
\centering
\includegraphics[width=0.95\columnwidth]{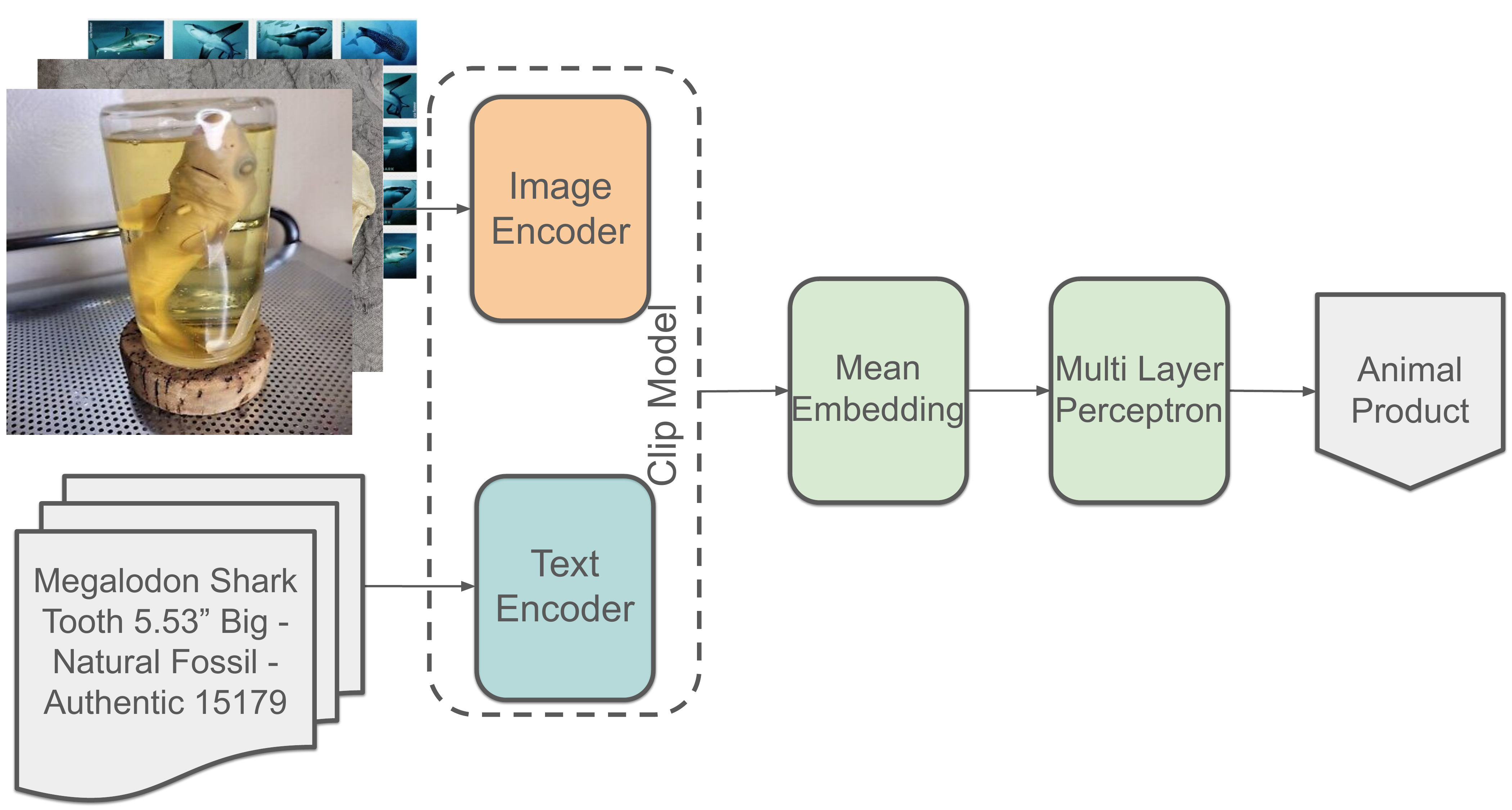}
% \vspace{-.3cm}
\caption{For the the multi-modal model, the average of image and text embeddings generated by the CLIP model is input into a multi-layer perceptron to generate the final classification of animal/not animal product.}
\Description{}
% \vspace{-.2cm}
\label{fig:multimodal}
\end{figure}

\myparagraph{Machine Learning Classifiers}
We use data derived using the three sampling strategies (Section~\ref{sec:training-data-generation}) for three use cases: leather products, shark products and \review{animal products.}
% In Section \ref{sec:use-cases}, we will apply these two models to filter irrelevant ads in our use cases: leather products and shark trade. 
%capable of identifying animal product ads in our data collection. The goal is to remove any possible ads that are irrelevant to our domain-specific analysis. 
%In our case, all ads that are not related to animal products or animals being sold online need to be removed and we keep the advertisements of animal by-products. Two models will be trained: a text model and and multi-modal model, using the images available in the ads.\\
% 
Since the ads contain both textual information and images, we train two distinct models for each use case: a text-based and a multi-modal model.

\noindent \textit{Text-Based Model.}
%In all three approaches 
%We use the text from the ads sampled as input to GPT-4 as shown in figure \ref{fig:Prompt}. 
We fine-tune a Bert-based model that uses the same input provided to GPT-4 (Figure~\ref{fig:Prompt}), using the Hugging Face \cite{huggingface@github} trainer class with an early stop to avoid overfitting:
% \jf{add ref to hugging face}
\setlist[enumerate]{leftmargin=5mm}

\begin{enumerate}
    \item \textit{Tokenizer.} We employ the BERT tokenizer from the Hugging Face Transformers library. This tokenizer is used to preprocess the input data, converting text into tokens suitable for input into the subsequent model layers.
    \item \textit{Model.} Our model architecture is based on the "bert-base-uncased" model for leather and sharks products, and it uses the \review{"bert-base-multilingual-cased" for animal products}, also from the Hugging Face Transformers library. We utilized the \textit{BertForSequenceClassification} class, specifying the number of labels=2 to indicate binary classification.
    \item \textit{Model arguments.} We utilize a strategy that evaluates at the end of each epoch, allowing for efficient monitoring  model performance. The early stop is applied with a patience of 5, which means the model will stop the training process when the validation loss increases, waiting for 5 more epochs and saving the best model in the end.
    \item \textit{Grid-Search.} We employ a Grid-search to find the optimal hyperparameters for the model to improve accuracy of predictions. For this model, we search for learning rate and weight decay.
    % \item \textit{Evaluation.} We evaluated the performance of our fine-tuned model using standard metrics such as accuracy, precision, recall, and F1-score on the validation dataset. 
\end{enumerate}

\setlist[enumerate]{leftmargin=5mm}

\noindent
\textit{Multi-Modal Model: Combining Image and Text.}
Since ads often include images, to explore the usefulness of these images for ad classification, we create a multi-modal model that combines text and images.  
%
% \myparagraph{Implementation details}
We apply the CLIP (Contrastive Language-Image Pretraining) model \cite{radford2021learning} to make use of transfer learning. 
% \jf{need reference for clip}
Our approach combines the embeddings of the image and text and passes the combined embedding through a multi-layer Perception as depicted in Figure~\ref{fig:multimodal}. The implementation details are as follows:

% \vspace{-.2cm}
\begin{enumerate}
\item \textit{Image and Text Embeddings.}
% The embeddings are created using the pre-trained model `clip-ViT-B-32'. This model is a pre-trained CLIP model implemented by SentenceTransformer \cite{reimers-2019-sentence-bert}. To combine both embeddings, we take the mean before passing to a Multi-Layer Perceptron.
We use the SentenceTransformer library's implementation of the CLIP model (`clip-ViT-B-32') to obtain embeddings \cite{reimers-2019-sentence-bert}, which are then combined by taking their mean before passing them to a Multi-Layer Perceptron.
% The implementation details are as follows:
% \jf{this seems broken. Where are the details?} I believe this is the sentence from the previous paragraph. should not be here.

\item \textit{Multi-Layer Perceptron.}
The model is a sequential network with three hidden layers. The first hidden layer consists of 512 units and uses the ReLU (Rectified Linear Unit) activation function. The second hidden layer comprises 256 units with ReLU activation. Each of these layers is followed by a dropout layer. Lastly, the output layer has 1 unit and uses the sigmoid activation function, since this is a binary classification task.

\item \textit{Grid-Search.} We employ a Grid-search to find the optimal hyperparameters for the model.
%to find the highest accuracy of predictions. 
We search for dropout rate, learning rate, optimizer, and weight decay.\\
\end{enumerate}

\vspace{-.4cm}
The model is trained using the Adam optimizer \cite{kingma2014adam}, a popular choice for its efficiency in handling various neural network architectures. We employ the binary cross-entropy loss function since we are working with a binary classification task. To monitor the model's performance and prevent overfitting, we utilize early stopping.
%with patience of 3 epochs.
This stops training if the validation loss fails to improve for \textit{n} consecutive epochs, indicating potential overfitting. Additionally, the best-performing model based on the validation loss was saved during training. This ensures access to the model with the optimal balance between training and generalization.

% an image based summary of the 3 approaches; key differences and expectations of the 3 approaches; is each approach improving upon some aspect of the previous one?

%% file: use-cases.tex
In this section, we discuss use cases that show the benefits of our approach in enabling the curation of wildlife advertisements at scale and helping experts uncover insights into the inner workings of online wildlife trafficking.

\subsection{Understanding Small Leather Products and Sellers on eBay}
\label{subsec:use-case1}

Since the COVID-19 pandemic, a major change in illegal wildlife trade has been observed, indicating a significant shift from offline to online marketplaces.
Moreover, this trade has been operated in plain sight, i.e., on the clear web and primarily on e-commerce platforms. 
Understanding such operations is essential for devising interventions and prevention mechanisms. The illegal trade of small leather products is a significant concern within the broader context of illegal trade and wildlife trafficking, as studies indicate that such products are among the most prevalent illegal wildlife products trafficked globally \cite{petrossian2016overview, van2019comparison}.

A small leather product typically refers to a variety of items made from animal skin. Small leather products made from exotic animal species are often associated with luxury, hence attracting high prices. The targeted species for small leather products vary in their protection around the globe, and some part of the trade of these products made from these species is permitted. However, open-web trade is largely unregulated, complicating the efforts of monitoring and control of these activities online. This has created a gray market that allows the sellers to launder forbidden products. Through these lenses, an attempt has been made to understand seller typologies for small leather products made from protected species,
taking into account a spectrum of products that range from legal to illegal and views such sales as potentially illegal. 

The main domain of this study focuses on eBay sellers. 
From those ads, only 15606 ads were classified as animal products, with a number of 8347 different sellers. We study the top 100 sellers who make 20 percent (3086) of the total number of sales.
The best model was used to classify the data collected and the animal ads were provided to \emph{criminologists} for further analysis.

\myparagraph{Seller Typologies} The results of the preliminary analysis indicate that 67\% (n=99) of sellers sell small leather products potentially made from protected species. The analysis of the seller profiles revealed two distinct typologies of sellers engaged in the trade of these products, based on the types and the volume of products sold by these sellers. 

\noindent \emph{Specialized sellers} deal with a specific type of product, such as leather skin, leather boots, watch accessories (leather straps), and Shamanic products potentially made from protected species. About 68\% (n=67) of sellers dealing in these small leather products fell within this category of sellers. 

\noindent \emph{Opportunistic sellers} deal with generic products, such as a variety of luxury goods and vintage wear. About 32\% (n=33) of sellers dealing in small leather products fell within this category of sellers. Opportunistic sellers do not specialize in any specific small leather product and do not seem to have the same volume or variety of products on their sites, and the listings that appear on their seller profiles appear to be random.

The typologies of the sellers extracted from the analysis of their profiles are consistent with those proposed by Dominguez et al (2024) \cite{dominguez2024online}, where the analysis of online illegal reptile trade in the Netherlands revealed three types of sellers: organized traders (which align more specifically with organized criminal groups, a distinction which was impossible to make in our study); professional traders (which, in our case, is the specialized seller); and enthusiasts (which, in our case, is the opportunistic seller). 

\begin{figure}[t] 
\centering
\includegraphics[width=0.9\columnwidth]
   {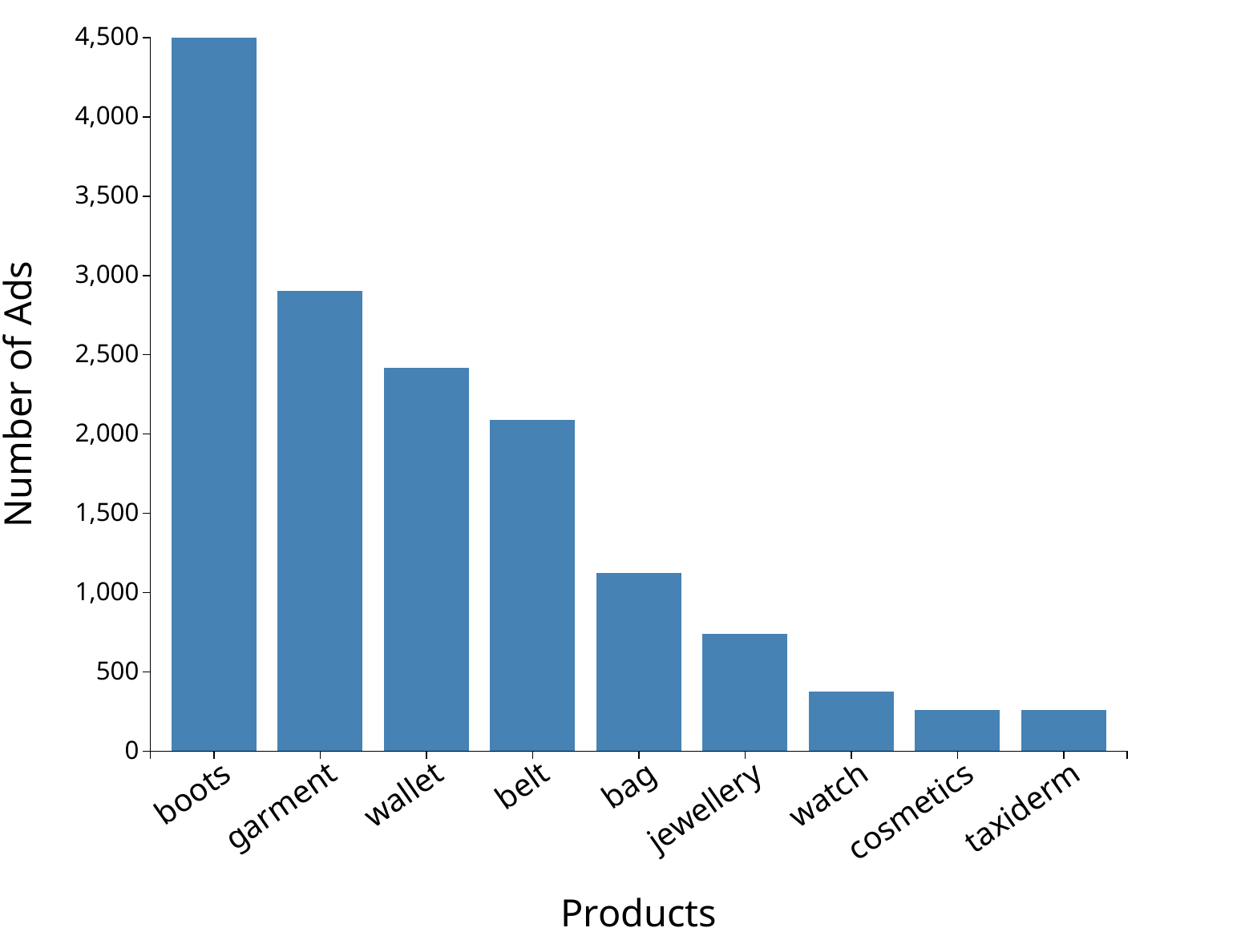}
    \label{fig:products}
    \caption{Distribution of ads for different products}
\end{figure}
\hfill
\begin{figure}[t]
    \centering
    \includegraphics[width=0.9\columnwidth]{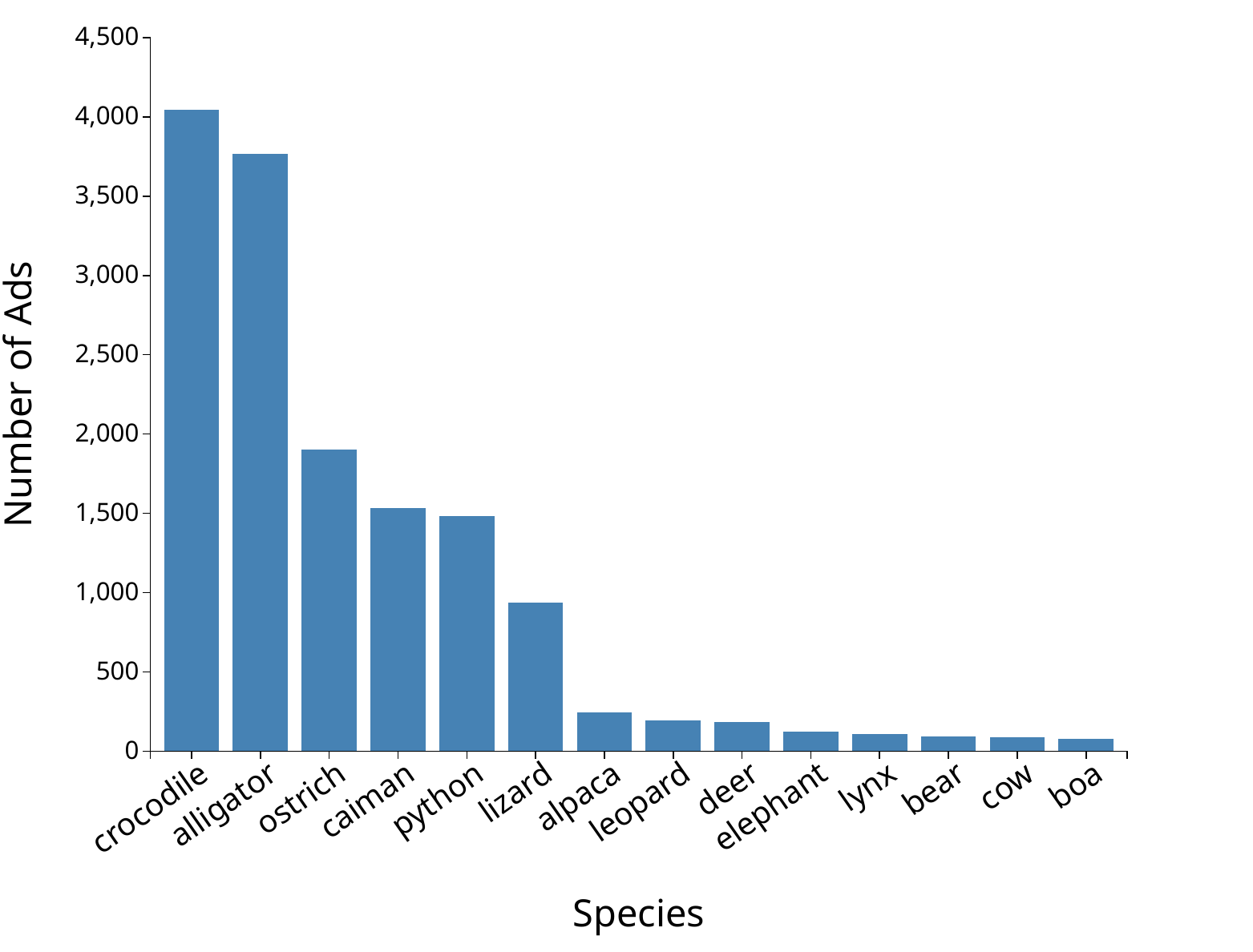}
    \label{fig:species}
\caption{Distribution of ads for different species }
\Description{}
\end{figure}

\myparagraph{Products, Species, and Locations} The most common small leather products available on eBay is boots with 4,500 ads, followed by clothes, bags, accessories, and taxidermy
(Figure \ref{fig:products}). This was similar regardless of which seller profile was examined. Products made from crocodile skin were the most frequent items, whereas other species included caimans, ostriches, pythons, deer, lizards, and stingrays (Figure \ref{fig:species}). The product description provided by the sellers often contained significant details about the product, including the species, the type of the product (if it is genuine or faux), brand name, dimensions, and its condition. The United States emerged as the top vendor country for these products, followed by Vietnam, the United Kingdom, and Thailand.

\subsection{The Online Trophy Trade of Sharks, Rays and Chimaeras}
\label{subsec:use-case2}

Environmental scientists want to better understand the trade in trophies from cartilaginous fishes or Chondrichthyes, a highly endangered clade of wildlife that includes sharks, rays, and chimaeras.
Chondrichtyan fishes are among the most imperiled groups of animals on the planet, with more than one-third of species classified as threatened with extinction \cite{dulvy2021overfishing}. Sharks are targeted for their fins, with estimates that between 23 and 73 million sharks are killed annually \cite{pytka2023internet}. Trade in shark fins and meat has been extensively studied, but very little is known about the trophy trade.

Quantifying and monitoring this trade is imperative, considering the high prices shark jaws can fetch in online marketplaces and the resulting incentive for their capture. This information can provide leads to determine the origin of chondrichthyan trophies, which may supplement demand streams for commercial fisheries or create unique threats from recreational fisheries such as land-based shark fishing, which is on the rise globally \cite{hingley2020conservation}.

We used our data curation pipeline and \ourapproach to create a multi-modal model to identify trophy trade of Chondrichtyan fishes (Section~\ref{sec:experimental-evaluation}). The collected ads were analyzed by environmental scientists who found that
restricted animals can be found on eBay. 
The analysis is described below.

\myparagraph{Restricted Animals on eBay}
Preliminary results from a complete crawl of eBay.com, one of the largest online marketplaces for shark and ray trophies, reveal that although the platform has introduced policies to restrict sharks listed on CITES Appendix II, their derivatives are still circulated on the platform.\footnote{\url{https://www.ebay.com/help/policies/prohibited-restricted-items/animal-products-policy?id=5046}} Jaws of requiem sharks (\textit{Carcharhinus spp.}) are still widely traded, including bull (\textit{C. leucas}), spinner (\textit{C. brevipinna}), and silky (\textit{C. falciformis}) sharks, which remain openly advertised in eBay listings. The most traffic, however, was observed in tiger shark (\textit{Galeocerdo cuvier}) jaws, comprising over one-fifth of common and species name matches and asking prices up to nearly \$3,000. High traffic was also observed in other unregulated species, including rare deep-sea species such as megamouth sharks (\textit{megachasma pelagios}) and spotted ratfish (\textit{hydrolagus colliei}), suggesting a pipeline from deep sea commercial fishing vessels to the trophy trade in the United States. Over 85\% of listings were linked to sellers in the United States.

%% file: conclusion.tex
In this paper we propose \ourapproach, a cost-effective approach that leverages Large Language Models to create effective classifiers that can be used to identify wildlife ads. \ourapproach streamlines the process required to create learning classifiers, enabling the study of different facets of the wildlife trade. We show through an experimental evaluation that \ourapproach creates diverse and representative sample, which when used for training, lead to classifiers that attain high F-measure. One surprising result of our evaluation was that the simpler classifiers outperformed
GPT-4 in one of the use cases and demonstrated comparable performance in  the others. This suggests that our approach can attain good classification performance at a low cost. Besides, to create the samples \ourapproach incurs a one-time cost, and the derived classifier can be applied multiple times. 

We used \ourapproach  to generate datasets that were analyzed by criminologists and environmental scientists. As we discussed in Section~\ref{sec:use-cases}, the data has led to 
new insights into different aspects of the wildlife trade. This demonstrates the usefulness of the approach and its potential to enable a deeper understanding of wildlife trafficking.

We designed our data gathering and curation pipeline so that it is extensible and can be customized for different tasks. 
In addition to wildlife trafficking, our approach can be potentially useful in other data triage tasks, such as identifying Web content related to different types of crime, including human trafficking and illegal gun sales~\cite{memex-site,memex-scientific-american}, which require the creation of learning-based classifiers. Our code is open-source \footnote{https://github.com/VIDA-NYU/LTS} and we hope
that its adoption by researchers working in the wildlife domain will encourage the community to contribute and share datasets, potentially enhancing the collective understanding of wildlife trafficking dynamics.

We have taken both ease of deployment and resource usage while choosing our target models. We used the text-only BERT fine-tuned, which is easy to deploy and cheap to run. The model can be hosted on platforms like Hugging Face with minimal setup and can be accessed via simple API calls for inference. The multi-modal (MM) model uses embeddings generated by the CLIP model (from Sentence-Transformers). With CLIP we can use a shared space for both image and text data, reducing the complexity and memory overhead associated with maintaining separate models for each modality. In our current configuration, a shared environment (a container in a Kubernetes cluster with access to 8 CPU cores of an Intel Xeon CPU E5-2640 v3 @ 2.60GHz), the model takes 32.64 minutes to process 15,379 inferences -- approximately 0.128 seconds per inference. This shows that, even without GPU acceleration, the model is capable of delivering reasonably fast inference times on a CPU.
There are additional avenues for improving performance, including batch processing, model quantization, or using GPU for optimized inference. These strategies can reduce both the computation time and memory footprint, making the multi-modal approach more scalable and practical for a variety of deployment environments.

Our results are promising and suggest several directions for future work.
Our approach is effective for domains (and research questions) for which LLMs have sufficient knowledge to perform classification. For domains in which this is not the case, it would be interesting to explore how to combine LLMs with other data labeling approaches, including data programming and crowdsourcing \cite{ratner2017snorkel, heo2020inspector, whang2023data}. 
An important goal of our project is to enable subject matter experts to gather data to answer their research questions. Therefore, we need easy-to-use interfaces for them to specify (and refine) their questions as well as the infrastructure to prepare the data and make them available in a database where they can be efficiently analyzed.  This involves challenges that lie at the intersection of data management and HCI.
Another direction for future research involves the addition of a more robust way to design the prompt used to explain the task to the LLM. There are some promising strategies for doing so including the use of dynamic examples generated with RAG and the use of a programmatic way to optimize the prompt such as DSpy~\cite{khattab2024dspy}.

Our work has enabled new research in environmental sciences and criminology. We hope that it can serve as an example for the data management community of how challenges in other domains can uncover new challenges in data management, and by addressing these challenges, we can both advance the state of the art and have practical impact.